%% file: arxiv_MaxEnt2024_paper.tex
\documentclass[11pt,pdftex]{article}
\usepackage[a4paper, total={6.8in, 8in}]{geometry}

\usepackage[utf8]{inputenc}
\usepackage{amsmath}
\usepackage{amsfonts}
\usepackage{amssymb}
\usepackage{cite}

\usepackage{graphicx}
\graphicspath{{Figs/}}

\usepackage{macros_gpi}
\input{block_defs.tex}

\input{NN_figures.tex}

\title{Model Based and Physics Informed Deep Learning Neural Network  Structures$^\dagger$}

\author{Ali Mohammad-Djafari $^{1,2}$ orcid number:{0000-0003-0678-7759}, \\ 
Ning Chu$^{2}$, Li Wang$^{3,4}$, Caifang Cai$^{2}$, and Liang Yu$^{5,6}$
\\ ~ \\ 
{\small \btabu{@{}l@{}}
$^1$ \quad International Science Consulting and Training (ISCT), 91440 Bures sur Yvette, France; djafari@ieee.org
\\
$^2$ \quad Zhejiang Shangfeng special blower company, Shaoxing 312352, China; chuning1983@sina.com
\\
$^3$ \quad Central South University, Changsha, China; li.wang.csu@csu.edu.cn
\\
$^4$ \quad Khalifa University, Abu Dhabi, UAE; 
\\
$^5$ \quad School of Civil Aviation, Northwestern Polytechnical Univ., Xian 710072, China; liang.yu@nwpu.edu.cn 
\\
$^6$ \quad State Key Laboratory of Airliner Integration Technology and Flight Simulation, Shanghai 200126, China. 
\\ ~\\ 
Presented at MaxEnt 2024:   
International Workshop on Bayesian Inference and Maximum Entropy \\ 
Methods in Science and Engineering, 
Gent university, Gent, Belgium, July 1-5, 2024.
\\ 
A modified version of this paper will appear in MaxEnt2024 Proceedings.
\etabu
}}

\begin{document}

\maketitle

\begin{abstract}
Neural Networks (NN) has been used in many areas with great success. When a NN's structure (Model) is given, during the training steps, the parameters of the model are determined using an appropriate criterion and an optimization algorithm (Training). Then, the trained model can be used for the prediction or inference step (Testing).
As there are also many hyperparameters, related to the optimization criteria and optimization algorithms, a validation step is necessary before its final use. 
One of the great difficulties is the choice of the NN's structure. Even if there are many "on the shelf" networks, selecting or proposing a new appropriate network for a given data, signal or image processing, is still an open problem. \\ 
In this work, we consider this problem using model based signal and image processing and inverse problems methods. We classify the methods in five classes, based on: i) Explicit analytical solutions, ii) Transform domain decomposition, iii) Operator Decomposition, iv) Optimization algorithms unfolding, and v) Physics Informed NN methods (PINN). Few examples in each category are explained.   
\\ ~\\ 
{\bf key words: } Deep Neural Network, Inverse problems; Bayesian inference; Model based DNN structure
\end{abstract}

\section{Introduction}
Neural Networks (NN), and in particular Deep Neural Networks (DNN) have been used in many area with great succes, mentioning here only a few main area: Computer vision, Speech recognition, Artificial Intelligence (AI) with Large Language Models (LLM). 

Nowadays, there are a great number of DNN models which are freely accessible for different use. However, when a NN's structure (Model) is selected, either, they are pre-trained and ready for use in that special domain of interest, or, we have selected one that we want to use for our proper application. In this case, in general, we have to follow the following steps: 
\ben
\item Training step: For this, we need to access the training data in a data base. Then, we have to choose an appropriate criterion, based on our category of the problem to be solved: Classification, Clustering, Regression, etc. The next step is to choose an appropriate optimization algorithm, some parameters such as the learning rate, and finally, train the model. This means that the parameters of the model are obtained at the end of the convergence of the optimization algorithm. 

\item Model validation and hyperparameters tuning: 
As there are many hyperparameters (related to the optimization criteria and optimization algorithms), a validation step is necessary before the final use of the model. In this step, in general, we may use a subset of the training data, or some other good quality training data, those which are qualified to be more confident, to tune the hyperparameters of the model and validate it. 

\item One last step before using the model is Testing it and evaluating its performances on a Testing data set. There are also a great number of testing and evaluation metrics, which have to be selected appropriately depending of the objectives of the considered problem.

\item Final step is uploading the implemented model and use it. The amount of the memory needed for this step has to be considered. 
\een

All these steps are very well studied and many solutions have been proposed. One of the great difficulties is, in fact, in the first step: the choice of the NN's structure. Even if there are many "on the shelf" networks, selecting or proposing a new appropriate network for a given data, signal or image processing, is still an open problem. Which structure to choose? How many layers are sufficient ? Which kind of layer is more effective: a dense one, a convolutional one?

In this paper, we consider this problem using model based signal and image processing and inverse problems methods. We classify the methods in five classes: 
\ben
\item Methods based on the Explicite analytical solutions;
\item Methods based on the Transform domain processing;
\item Methods based on Operator decomposition structure; 
\item Methods based on the Unfolding of the iterative optimization algorithms; and
\item Methods known as the Physics Informed NN (PINN).
\een
Few examples in each category are explained.    

The structure of the rest of this paper is the following: 
As we focus on Forward modeling and Inverse problems methods, in Section~2, we present first two  inverse problems that we will use later as the applications of the developped methods: Infrared imaging (IR), and Acoustical imaging. 
In Section~3, a short presentation of the Bayesian inference framework, and its link with the deterministic regularization methods, are presented.  
From this point, the four categories of the previously mentionned methods for selecting the structure of the NN are presented with details in the Sections 4, 5, 6  and 7. Finally, a very short applications of these methods in industry is presented  and conclusions of this work are given. 

\section{Inverse problems considered}
\subsection{Infrared Imaging}
In many industrial application, Infrared cameras are used to measure the temperature remotely. The the thermal radiation is propagated from the source to the front of the camera, and then transformed to a current measured by the CCDs of the infrared camera. If we represent the radiated source temperature distribution as $f(x,y)$, and the measured infrared energy distribution by  $g(x,y)$, a very simplified forward model relating them can be given as: 
\beq
\bg(x,y) = \intd h(x,y)*\phi(\rf(x,y)) \d{x}\d{y}
\label{eq01}
\eeq
In this very simplified model, $h(x,y)$ is a convolution kernel function, also called the Point spread function (PSF), approximating the propagation of the thermal diffusion, and  $\phi(.)$ is a nonlinear function which is, given by: 
\beq
\phi(f)=\phi_\theta(f)=\left[\frac{e f^n + (1-e)T_u^n + (\expf{-k d}-1) T_a}{\expf{k d}}\right]^{1/n},
\label{eq02}
\eeq
where $\theta=\{e,T_u,k,d,T_a\}$ is a set of parameters:  $e$ is the emissivity, $T_u$ is the background temperature, $k$ is the attenuation coefficient, which can be a function of the humidity, $d$ is the distance and $T_a$ is the air temperature \cite{udayakumar2013infrared,bhowmik2012advances,klein2016advanced,LiWang2022}. 

The convolution kernel can be obtained experimentally or assumed given, for example to be a Gaussian shaped with known or unknown width parameter. Figure~\ref{fig01} shows the forward model.
\bfig[htb]
\[
\btabu{@{}c@{}} $\rfxy$ \\ \includegraphics[scale=.2]{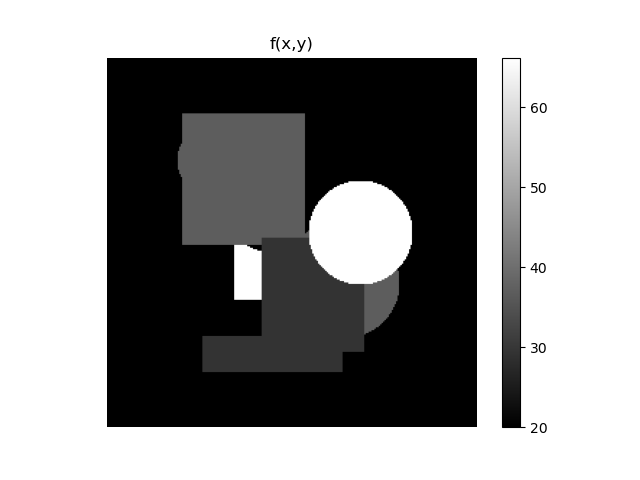} \etabu
\hspace{-2mm}\ra
\fbox{\btabu{@{}c@{}} Emissivity \\ environment \\ Simulator \\ $\phi(f)$ \etabu}
\ra
\btabu{@{}c@{}} $\phi(x,y)$ \\ \includegraphics[scale=.2]{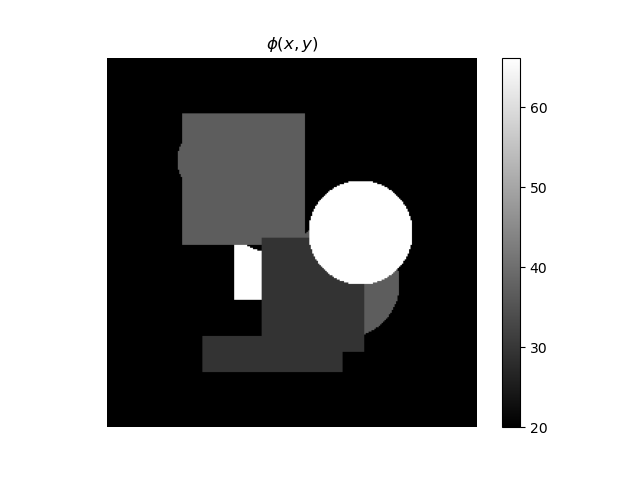} \etabu
\hspace{-2mm}\ra
\fbox{\btabu{@{}c@{}} Diffusion \\ Convolution \\ Simulator \\ $g=h*\phi$ \etabu}
\ra 
\btabu{@{}c@{}} $\bgxy$ \\ \includegraphics[scale=.2]{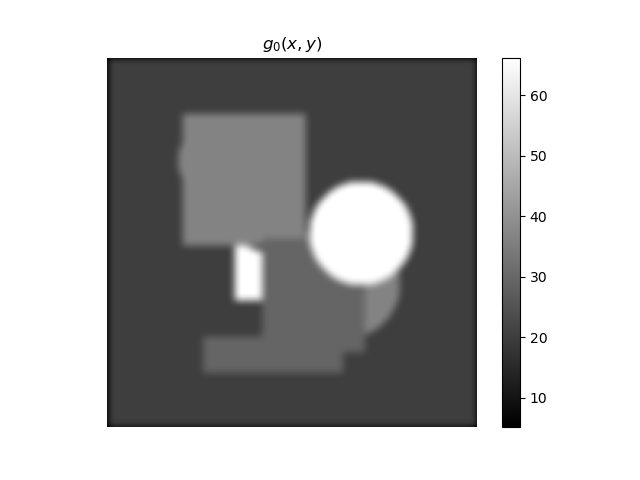} \etabu
\]
\caption{Infrared simplified Forward model: Real temperature distribution as input $\fxy$, nonlinear emissivity and environment perturbation function $\phi(f)$ and the  convolution operation to simulate the diffusion process, and finally the measured infrared camera output $\gxy$.}
\label{fig01}
\efig

Figure~\ref{fig02} shows the corresponding inverse problem. The solution is  
either obtained by inversion or by a Neural Network.   

\bfig[htb]
\bcc
\btabu{@{}c@{}c@{}c@{}}
IR image  $\bgxy$ & \btabu{@{}c@{}} Forward model  \\ $\Longleftarrow$ \etabu &  Actual temperature $\rfxy$
\\[-3pt]
\btabu{@{}c@{}} \includegraphics[height=4.2cm]{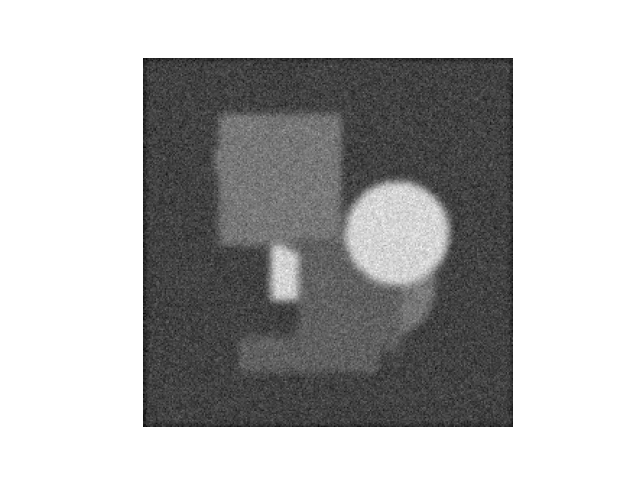}\etabu\hspace{-5mm}
& 
\btabu{@{}c@{}} \includegraphics[height=3.4cm]{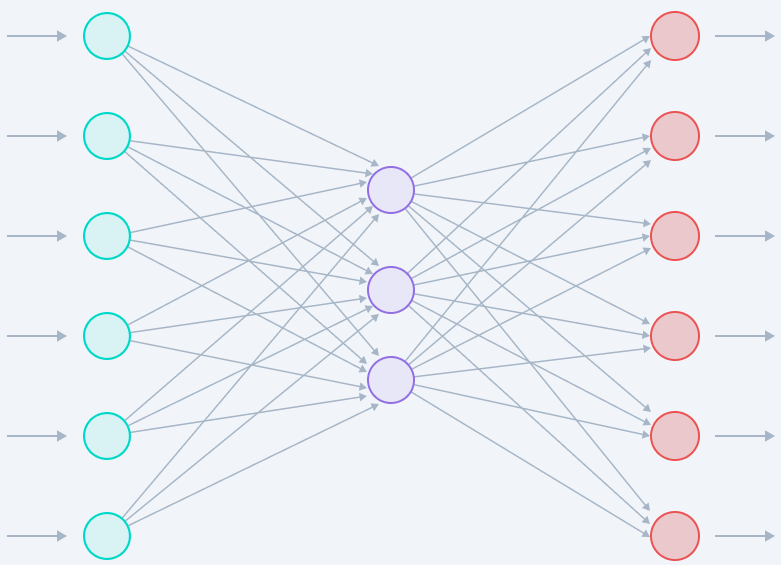} \etabu 
%\btabu{@{}c@{}} \NNg{4}{4}{4} \etabu
&
%\hspace{-5mm} 
\btabu{@{}c@{}} \includegraphics[height=4.2cm]{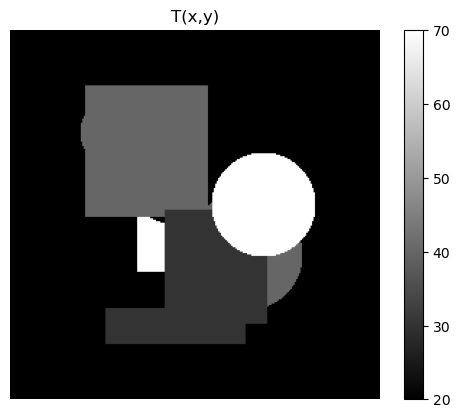}\etabu
\\[-3pt]
Data $\bg(x,y)$ & \btabu{@{}c@{}} Inversion or Neural Network  \\ $\Longrightarrow$ \etabu & $\rf(x,y)$
\etabu
\caption{Infrared imaging: Forward model and Inverse problem resolved either by mathematical inversion or by a Neural Network}
\label{fig02}
\ecc
\efig

\clearpage\newpage
\subsection{Acoustical Imaging}
In acoustical imaging we consider, a network of microphones receive the signals emitted by the sources with different delays:
\beq
\bg(x',y',t) = \int\d{\omega}\intd \rfxy \expf{-j\omega(t-\tau(x'-x,y'-y))} \d{x}\d{y},
\label{eq03}
\eeq
where $\rfxy$ is the intensity of the sources distribution at positions $(x,y)$, 
$\bg(x',y',t)$ is the received signal at positions $(x',y')$ of the microphones, $\omega$ is the frequency of the emitting source, and $\tau(x'-x,y'-y)$ is the the delai between the emitted source component at position $(x,y)$ to the microphone position $(x',y')$ of the microphones, which is a function of the sound speed and the distance $d=\sqrt{(x'-x)^2+(y'-y)^2}$. 
See \cite{Chu_2020Fast,CHEN2024111130,LUO2024111476} for more details.  
Figure~\ref{fig03} shows the forward model.  

\bfig[htb]
\bcc
\btabu{@{}c@{}c@{}c@{}c}
\\
 & ~~~~~~~~ Forward Model ~~~~~~~~ & & 
\\
\includegraphics[height=3cm,width=4cm]{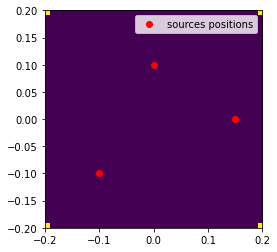}
& &
\includegraphics[height=3cm,width=3.5cm]{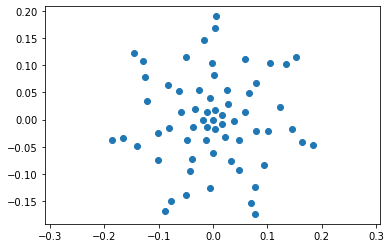}
& 
\includegraphics[height=3cm,width=3cm]{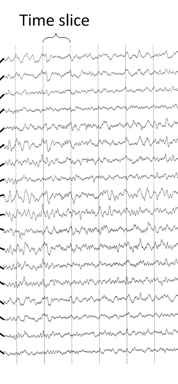}
\\
Sound sources $\fxy$ & & Microphones positions & Received signals $\bg(x',y',t)$ 
\etabu
\caption{Forward model in acoustical imaging: Each microphone receives the sum of the delayed sources sounds.}
\label{fig03}
\ecc
\efig
In a classical acoustical imaging, the inversion is made in two steps: Beamforming (BF) and Deconvolution. In BF, the received signals are delayed in opposite directions using the positions of the microphones and thus forming an image $b(x,y)$ which is showed that it is related to the actual sound source distribution $\rfxy$ via a convolution operation. Thus, the inverse problem is composed of two steps: Beamforming and Deconvolution. This last step can be down either in classical inversion or via a Neural Network \cite{Davis2020,Nelson2021,Roberts2022,sun2019deep,wang2021inverse,cheng2020end,cai2019deep}. Figure~\ref{fig04} shows these steps. 

\bfig[htb]
\bcc
\btabu{@{}c@{}c@{}c@{}c@{}c}
\includegraphics[height=3cm,width=3.2cm]{microphone_array_001}
&
\includegraphics[height=3.2cm,width=2cm]{microphone_signals}
&
\hspace{-2mm}\includegraphics[height=3cm,width=3.5cm]{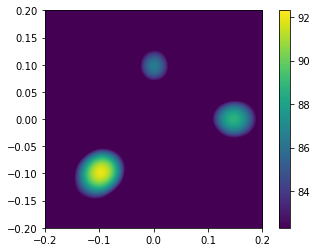}
& 
\includegraphics[height=2.8cm,width=2.5cm]{input-hidden-output-layers2}
&
\includegraphics[height=3cm,width=3.5cm]{source_positions}
\\
Data $\bg(x',y',t)$ & Beamforming (BF)   & $\green{b}(x,y)$ & {\bf Deconvolution} & $\rfxy$
\etabu
\caption{Acoustical imaging Inverse problem via Beamforming (BF) and Deconvolution (Dec):: In a first step, the received data are used to obtain an image $\green{b}(x,y)$ by BF, then an inversion by Dec can result to a good estimate of the sources.}
\label{fig04}
\ecc
\efig

\section{Bayesian Inference for Inverse Problems}
As the main subject of this paper is to explain the link between the inverse problems and Neural Networks, and as the Bayesian inference approach is an appropriate one for defining solutions for them, in this section, very fast, we summarize the Bayesian approach, specifically for inverse problems \cite{AMD2021}. Starting by the simple case of linear inverse problems $\bgb=\Hb\rfb+\epsilonb$ and Gaussian prior models for the unknown image $\rfb$ and the observation noise $\epsilonb$, we easily can easily summarize this approach in the following Figure~\ref{fig05}:  

\bfig[htb]
\bcc
\btabu[b]{c}
\begin{tikzpicture}[node distance=1cm]
\node (vf) [startstop] {$\bvf$};
\node (veps) [startstop, xshift=2cm] {$\bve$};
\node (f) [process, below of=vf] {$\rfb$};
\node (g) [process, below of=f] {$\bgb$};
\draw [arrow](vf) -- (f);
\draw [arrow](veps) -- (g);
\draw [arrow](f) -- (g);
\node (H) [note, below of=f,xshift=-.3cm,yshift=5mm] {$\Hb$};
\node (pgf) [note, below of=f,xshift=5cm] {$p(\bgb|\rfb,\bve)=\Nc(\bgb|\Hb\rfb,\bve\Ib)\propto\expf{-\frac{1}{2\bve}\|\bgb-\Hb\rfb\|^2}$};
\node (pf) [note, below of=veps,xshift=3cm] {$p(\rfb|\bvf)=\Nc(\rfb|\zerob,\bvf\Ib)\propto\expf{-\frac{1}{2\bvf}\|\rfb\|^2}$};
\node (pfg) [note, below of=g,xshift=2cm] {Bayes: $p(\rfb|\bgb,\bvf,\bve)\propto p(\bgb|\rfb,\bve)p(\rfb|\bvf)\propto\expf{-\frac{1}{2\bve}\|\bgb-\Hb\rfb\|^2-\frac{1}{2\bvf}\|\rfb\|^2} =\Nc(\rfb|\rfbh,\rSigmabh)$};
\end{tikzpicture}
\etabu
\caption{Basic Bayesian approach illustration for the case of Gaussian priors.}
\label{fig05}
\ecc
\efig
Then, we can easily obtain the following relations:
\beq
p(\rfb|\bgb,\bvf,\bve)=\Nc(\rfb|\rfbh,\rSigmabh), \mbox{~with~}
\left\{\barr{l}
\rfbh=[\Hb'\Hb+\lambda\Ib]^{-1}\Hb'\bgb\\
\rSigmabh=\bve[\Hb'\Hb+\lambda\Ib]^{-1}, \quad \lambda=\frac{\bve}{\bvf}
\earr\right..
\label{eq04}
\eeq
In this particular Gaussian case, as the expected value and the mode are the same, the computation of $\rfbh$ can be done by an optimization, the Maximum A Posteriori (MAP) solution: 
\beq
\rfbh=\argmax{\rfb}{p(\rfb|\bgb,\bvf,\bve)}=\argmin{\rfb}{J(\rfb)}
\mbox{~where~}
J(\rfb)=\|\bgb-\Hb\rfb\|_2^2+\lambda\|\rfb\|_2^2, \quad \lambda=\frac{\bve}{\bvf}.
\label{eq05}
\eeq
However, the computation of $\rSigmabh$ is much more costly. There are many approximation methods such as the MCMC sampling methods, the Perturbation-Optimization, Langevin sampling, or Variational Bayesian Approximation (VBA) \cite{Foxtutorial}.

\section{NN Structures based on Analytical solutions}
\subsection{Linear analytical solution}
As we could see in the previous section, the solution of the inverse problem, in that particular case of linear forward model and Gaussian priors, is given by: 
\beq
\rfbh=(\Hb\Hb^t+\lambda \Ib)^{-1}\Hb^t \bgb=\Ab\bgb=\Bb \Hb^t \bgb=\Hb^t(\frac{1}{\lambda}\Hb\Hb^t+\Ib)^{-1})\bgb=\Hb^t\Cb\bgb.
\eeq
These relations are shown schematically in Figure~\ref{fig06}. 
\bfig[hbt]
\[
\bgb\ra\fbox{$~~\Ab~~$ }\ra\rfbh 
\quad \mbox{or}\quad 
\bgb\ra\fbox{$~~\Hb^t~~$}\ra\fbox{$~~\Bb~~$ }\ra\rfbh
\quad \mbox{or}\quad 
\bgb\ra\fbox{$~~\Cb~~$}\ra\fbox{$~~\Hb^t~~$ }\ra\rfbh
\]
\caption{Three NN structures based on analytical solution expressions.}
\label{fig06}
\efig

\bit
\item $\Hb^t$ can be implemented directly or by a Neural Net (NN)

\item When $\Hb$ is a convolution operator, $\Hb^t$ is also a convolution operator and can be implemented by a Convolutional Neural Net (CNN).

\item $\Bb=(\Hb\Hb^t+\lambda \Ib)^{-1}$ and $\Cb=(\frac{1}{\lambda}\Hb\Hb^t+\Ib)^{-1})$ are, in general, dense NNs.

\item When $\Hb$ is a convolution operator, $\Bb$ and $\Cb$ can also be approximated by convolution operators, and so by CNNs \cite{mohammaddjafari2023deeplearninginverseproblems,mohammaddjafari2023deeplearningbayesianinference,niven2024dynamicalidentificationmodelselection}. 
\eit
%---------------------------------------------------------------
\subsection{Feed Forward and Residual NN Structure}
Let us to consider the case of denoising problem: 
\(
\bgb=\rfb+\repsilonb, 
\)
where $\rfb$ is the original, $\repsilonb$ is the noise and $\bgb$ the observed noisy image. Let us also to consider the case of linear denoising method, obtained by a one layer dense Feed-Forward network: \quad  $\rfbh=\Ab(\bgb)$, shown in Figure~\ref{fig07}. 
\bfig[htb]
\[
\bgb\lra\fbox{$~~~~\Ab~~~~$}\lra\rfbh
\]
\vspace{-12pt}
\caption{Feed Forward NN structure.}
\label{fig07}
\efig

If, in place of directly searching for the denoised solution, first use a NN to obtain the noise and then, obtain the denoised image by substraction: 
$\repsilonbh=\Ab(\bgb)\ra\rfbh=\bgb-\repsilonbh$, we obtain the Residual Network structure, shown in Figure~\ref{fig08}.
 
\bfig[hbt]
\bcc
\begin{tikzpicture}[node distance=1cm]
\node (NI) [note] {$\bgb$};
\node (RA1) [note, xshift=1cm] {$.$};
\node (ADD) [process, xshift=3.5cm] {+};
\node (RA2) [note, below of=RA1] {$.$};
\node (RA3) [note, below of=ADD] {$.$};
\node (RA4) [note, xshift=4.2cm] {$\ra$};
\node (RA5) [note, below of=RA4] {$\ra$};
\node (DI) [note, xshift=4.7cm] {$\rfbh$};
\node (EN) [note, below of=DI] {$\repsilonbh$};
\node (ED) [process, below of=NI, xshift=2.3cm] {$\Ab$};
\draw [arrow](NI) -- (RA1);
\draw [arrow](RA1) -- (RA2);
\draw [arrow](RA2) -- (ED);
\draw [arrow](ED) -- (RA3);
\draw [arrow](RA3) -- (ADD);
\draw [arrow](RA1) -- (ADD);
\draw [arrow](RA3) -- (RA5);
\end{tikzpicture}
\ecc
\vspace{-12pt}
\caption{Residual structure NN.}
\label{fig08}
\efig

\newpage
\section{NN Structures based on Transformation and Decomposition}
When the inverse operator $\Ab$ exists and is a linear operation, it can always be decomposed by three operations; Transformation $\Tb$, Transform domain operation $\Ob$, and Inverse Transform $\Tb^{-1}$, or Adjoint Transform $\Tb^{T}$. This structure is shown in Figure~\ref{fig09}.: 

\bfig[htb]
\[
\bgb\lra\fbox{$~~~~\Ab~~~~$}\lra\rfbh 
\qquad \Lra \qquad 
\bgb\lra\fbox{$~~\Tb~~$}\lra\fbox{$~~\Ob~~$}\lra\fbox{$~~\Tb^T~~$}\lra\rfbh
\]
\caption{Transform domain NN structure.}
\label{fig09}
\efig

Two such transformations are very important: Fourier Transform (FT) and Wavelet Transfor (WT).  
%---------------------------------------------------------------
\subsection{Fourier Transform based Networks}

FT-Drop Out-IFFT: \qquad 
\(
\rfbh=\Fb^T\left[\Lambdab\Fb[\bgb]\right]
\), 
which results to the FTNN shown in Figure~\ref{fig10}.
\bfig[htb]
\[
\bgb\lra\fbox{~~FFT~~}\lra\fbox{~~DO~~}\lra\fbox{~~IFFT~~}\lra\rfbh
\]
\vspace{-12pt}
\caption{Fourier Transform domain NN structure.}
\label{fig10}
\efig

%---------------------------------------------------------------
\subsection{Wavelet Transform based Networks}

Wavelet Transform-Drop Out-Inverse Wavelet Transform is defined as: 
\(
\rfbh=\Wb^T\left[\Lambdab\Wb[\bgb]\right]
\),
which results to the WTNN shown in Figure~\ref{fig11}.

\bfig[h!]
\[
\bgb\lra\fbox{~~ WT ~~}\lra\fbox{~~DO~~}\lra\fbox{~~ IWT ~~}\lra\rfbh
\]
\vspace{-12pt}
\caption{Wavelet Transform domain NN structure.}
\label{fig11}
\efig
We may note that, for image processing, in FT, there is one chanel, and in WT, there are 4 or more channels.

%---------------------------------------------------------------
\subsection{NN Structures based on operator decomposition and Encoder-Decoder Structure}
A more general operator decomposition can be: 
\bcc
Encoder-Decoder: \qquad $\Ab(\bgb)=\Ab_{Dec}(\Ab_{Enc}(\bgb))$
\qquad 
$\bgb\ra\fbox{$\Ab_{Enc}$}\ra\fbox{$\Ab_{Dec}$}\ra\rfbh$ 
\ecc
Each part, Encoder and Decoder, can also be decomposed on a serie of partial operators: 
\bcc
\btabu{cc}
\btabu{c}
$\Ab_{Enc}(\bgb)$: $N$ layers Network: \\
$\Cb_0=\Eb_0(\bgb)$, \\ 
$\Cb_1=\Eb_1(\Cb_0)$, \\ 
$\Cb_2=\Eb_2(\Cb_1)$, \\ 
$\Cb_{N-1}=\Eb_N(\Cb_N)$,\\ 
$\vdots$, \\ 
$\Ab_{Dec}(\bgb)=\Cb_{N-1}$ 
\etabu
&
\btabu{c} 
$\Ab_{Dec}(\bgb)$: $N$ layers Network: \\
$\bar{\Cb}_{N-2}=\Db_{N-1}(\Cb_{N-1})$, \\ 
$\vdots$, \\ 
$\bar{\Cb}_2=\Db_3(\bar{\Cb}_3)$, \\ 
$\bar{\Cb}_1=\Db_2(\bar{\Cb}_2)$, \\ 
$\bar{\Cb}_0=\Db_1(\bar{\Cb}_1)$, \\ 
$\Ab(\bgb)=\bar{\Cb}_0=\rfbh$
\etabu
\etabu
\ecc
which results to the Encoder-Decoder NN shown in Figure~\ref{fig12}.

\bfig[h]
\bcc 
$\bgb\ra$\fbox{$\Eb_0$}$\ra\Cb_0\ra$\fbox{$\Eb_1$}$\ra\Cb_1\ra$\fbox{$\Eb_2$}$\ra\Cb_2\ra\cdots\ra\Cb_{N-1}$ 
\\ 
~\qquad $\Cb_{N-1}\ra$\fbox{$\Db_{N-1}$}$\ra\Cb_{N-2}\ra\cdots\ra\bar{\Cb}_1\ra\fbox{$\Db_1$}\ra\bar{\Cb}_0=\rfbh$
\ecc
\vspace{-12pt}
\caption{Encoder Decoder structure based on Operator decomposition.}
\label{fig12}
\efig

%---------------------------------------------------------------
\section{DNN structures obtained by unfolding optimization algorithms}

Considering linear inverse problems: $\bgb=\Hb\rfb+\epsilonb$ within the Bayesian framework, with Gaussian errors, double exponential prior, and MAP estimation, or equivalently, the $\ell_1$ regularization framework, the inversion becomes the optimization of the following criterion:
\beq
J(\rfb) = \|\bgb-\Hb\rfb\|_2^2 + \lambda \|\rfb\|_1
\eeq
which can be obtained via an iterative optimization algorithm, such as ISTA:  
\[
\hspace{-1cm}\rfb^{(k+1)} = 
Prox_{\ell_1} \left(\rfb^{(k)},\lambda\right) 
\defined 
\Sc_{\lambda/\alpha}\left(\alpha
\Hb^t\bgb + (\Ib - \alpha \Hb^t\Hb) \rfb^{(k)}\right)
\]
where $\Sc_{\theta}$ is a soft thresholding (ST) operator and  
$\alpha \le eig(\Hb^t\Hb)$ is the Lipschitz constant of the normal operator. 
Then, $(\Ib - \alpha \Hb^t\Hb) \rfb^{(k)}$ 
can be considered as a filtering operator,
$\alpha\Hb^t\bgb$ can be considered as a bias term, and 
$\Sc_{\theta}$ as a pointwize nonlinear operator. Thus, we obtain the unfolded structure shown in Figure~\ref{fig13}. 
\bfig[h]
\bcc
\Blocka
\ecc
\vspace{-12pt}
\caption{One iteration of an $\ell_1$ regularization optimization algorithm.}
\label{fig13}
\efig

Now, if we consider a finite number of iterations, we can create a Deep Learning network structure, shown in Figure~\ref{fig14}.

\bfig[h]
\bcc
{\small \BlockDLw}
\ecc
\vspace{-12pt}
\caption{DLL structure based on a few iterations of the regularization optimization algorithm.}
\label{fig14}
\efig

where $\rWb_0=\alpha\Hb$ and  
$\rWb^{(k)}=\rWb=(\Ib-\alpha\Hb^t\Hb), \quad k=1,\cdots,K$.  
A more robust, but more costly is to learn all the layers $\rWb^{(k)}=(\Ib-\alpha^{(k)}\Hb^t\Hb), \quad k=1,\cdots,K$.

\section{Physics Informed Neural Networks (PINN)}
The main idea behind the PINN are: 

\bit
\item NNs are universal {function approximators}. Therefore a NN, provided that it is deep enough, can approximate any function, and also the solution for the differential equations.

\item Computing the derivatives of a NN output with respect to any of its input (and the model parameters during backpropagation), using {Automatic Differentiation} (AD), is easy and cheap. This is actually what made NNs so efficient and successful.

\item Usually NNs are trained to fit the data, but do not care from where come those data. This is where physics informed comes into play. 

\item Physics based or Physics Informed: If, {besides fitting the data, fit also the equations that govern that system and produce those data}, their predictions will be much more precise and will generalize much better.
\eit
In the following, we consider two cases: i) the case where the forward model is explicitely given and an operator $\Hb$, ii) the cases where the forward model are given either by an ODE or PDE.

When the forward operator $\Hb$ is known, we can use it to propose the following structure, to define a criterion to use for optimizing the parameters of the NN. Main references on PINN are: 
\cite{raissi2019physics,lu2021deepxde,karniadakis2021physics,chen2020physics}.

\subsection{PINN when an explicie forward model is available}
When an explicie forward model $\Hb$ is available, we can use it at the output of the NN, shown in Figure~\ref{fig15}.
\bfig[h]
\[
\mbox{Data~}\bgb\lra\fbox{~~NN(\rwb)~~}\lra \rfbh\lra\fbox{\Hb}\lra\gbh(\rwb)\lra 
J(\bgb,\gbh(\rwb))=\|\bgb-\gbh(\rwb)\|^2
\]
\vspace{-12pt}
\caption{PINN when an explicie forward model $\Hb$ is available.}
\label{fig15}
\efig

Here, the NN is used as a proxi or surrogate inversion method which parameters $\wb$ are obtained by the optimization of the loss function $J(\bgb,\gbh)$. If for an experiment, we have a set of data, we can use them, to define a loss function with two parts, shown in Figure~\ref{fig16}.
\bfig[h]
\[
\mbox{Data~}\gb_i\lra\fbox{~~NN(\rwb)~~}\lra \rfbh_i\lra\fbox{\Hb}\lra\gbh_i\lra 
J(\bgb,\gbh(\rwb))=\frac{1}{\sigmae^2}\sum_i\|\bgb_i-\gbh_i(\rwb)\|^2+\frac{1}{\sigma_f^2}\sum_i \|\rfb_i-\bar{\rfb}\|^2
\]
\vspace{-12pt}
\caption{PINN when an explicie forward model $\Hb$ is available}
\label{fig16}
\efig

which can also be interpreted in a Bayesian way: 
\begin{align}
p(\bgb|\rfb,\sigmae^2)&=\Nc(\bgb|\Hb\rfb,\Sigmabe), \quad\mbox{with~}  \Sigmabe=\sigmae^2\Ib \\
p(\rfb|\sigmae^2)&=\Nc(\rfb|\bar{\rfb},\sigma_f^2\Ib), \quad\mbox{with~} \Sigmab_f=\sigma_f^2\Ib
\end{align}
One last extension is to define the loss function as: 
\beq
J(\bgb,\gbh)=\Tr{\Sigmabe}+\Tr{\Sigmab_f} \mbox{~~with~~} 
\Sigmabe=\sum_i [\bgb_i-\gbh_i][\bgb_i-\gbh_i]^t, \mbox{~~and~~} 
\Sigmab_f=\sum_i [\rfb_i-\bar{\rfb}][\rfb_i-\bar{\rfb}]^t.
\eeq
In this way, at the end of training, we also have a good estimate of the covariances 
$\Sigmabe$ and $\Sigmab_f$, which can, eventually, used as uncertainty priors for a new test input $\bgb_j$. 

\subsection{PINN when the forward model is described by an ODE or a PDE}

PINN for inverse problems modelled by ODE or PDE was conceptually proposed by Raissi et al., back in 2017\cite{raissi2019physics}. Since that, there have been many works on the subject. See for example  
\cite{yang2021b,zhu2019physics,yang2019adversarial,geneva2020modeling} and 
\cite{gao2021phygeonet,cai2021physics}. 

The main idea consists in creating a hybrid model where both the observational data and the known physical knowledge are present in model training. 

Inverse problems in ODE or PDE can be categorized in three, more and more difficult,  cases. To illustrate this, let us to consider a simple dynamical system: 
\bit
\item Unknown parameters model. As a simple dynamical system, consider: 
\[
\dpdx{u}{t} = \red{a} u + \red{b}, 
\]
where the problem becomes to estimate the the parameters $\rthetab=(\red{a},\red{b})$.
\item Unknown right side function: 
\[
\dpdx{u}{t} = \red{f(t,u)}.
\]
\item Unknown parameters as well as the unknown function $\red{f(t,u)}$: 
\[
\dpdx{u}{t} = \red{a} u + \red{b} + \red{f(t,u)}
\]
\eit
The difference between the classical NN and Physics informed NN for a dynamical system, such as $\dpdx{u}{t} = f(t,u)$, can be illustrated as follows.

\smallskip\noindent{\bf Classical NN:} \\ 
The criterion to optimize is just a function of the output error, shown in Figure~\ref{fig17}.
\bfig[h]
\[
t
\barr{c}
\NNsOneTwoTwoOne 
\earr
\hspace{-1.2cm}\rwb\hspace{1cm}
\hat{u}
\ra
J(\rwb)=\sum_t [u(t)-\hat{u}(t)]^2
\]
\vspace{-12pt}
\caption{Classical NN: The criterion to optimize to obtain the NN parameters $\rwb$ is just a function of the output error.}
\label{fig17}
\efig

\smallskip\noindent{\bf PINN:}\\ 
The criterion to optimize is a function of the output error and the forward error, shown in Figure~\ref{fig18}.
\bfig[h]
\[
t
\barr{c}
\NNsOneTwoTwoOne
\earr
\hspace{-1.2cm}\rwb\hspace{1cm}
\hat{u}
\ra
J(\rwb)=\sum_t [(u(t)-\hat{u}(t)]^2+\sum_t \left[\dpdx{u}{t}-f(t,u)\right]^2
\]
\vspace{-12pt}
\caption{PINN: The criterion to optimize to obtain the NN parameters $\rwb$ is a function of the output error and the physic based forward model error.}
\label{fig18}
\efig

\smallskip\noindent{\bf Parametric PINN:}\\ 
When there are some parameters $\rtheta$ of the PDE is also unknown, we may also estimate it at the same time with the parameters $\rwb$ of the NN, as shown in Figure~\ref{fig19}. 
\bfig[h]
\[
\barr{c}
t \\[12pt] \rtheta
\earr
\barr{c}
\NNsTwoTwoTwoOne
\earr
\hspace{-1.2cm}\rwb\hspace{1cm}
\hat{u}
\ra
J(\rwb,\rtheta)=\sum_t [(u(t)-\hat{u}(t)]^2+\sum_t \left[\dpdx{u}{t}-f_{\rtheta}(t,u)\right]^2
\]
\vspace{-12pt}
\caption{Parametric PINN: The unknown parameter $\rtheta$ of the PDE model can be optimized at the same time as the NN parameters $\rwb$.}
\label{fig19}
\efig

\smallskip\noindent{\bf Non parametric PINN Inverse problem:}\\ 
When the right hand side of the ODE is an unknown function to be estimated too, we may consider a separate NN for it, as shown in Figure~\ref{fig20}.
\bfig[h!]
\[
\barr{l}
t
\barr{c}
\NNsOneTwoTwoOne
\earr
\hspace{-1.2cm}\rwb_1\hspace{1cm}
\hat{u}
\\   
\barr{@{}c@{}}
u \\[12pt] t
\earr
\barr{c}
\NNsTwoTwoTwoOne
\earr
\hspace{-1.2cm}\rwb_2\hspace{1cm}
\rfh
\earr
\Ra
J(\rwb_1,\rwb_2)=\sum_t [(u(t)-\hat{u}(t)]^2+\sum_t \left[\dpdx{u}{t}-\rfh(t,u)\right]^2
\]
\vspace{-12pt}
\caption{Non parametric PINN: The unknown functio $\red{f}$ of the PDE model can also be approximated via a second NN, and the parameters $\rwb_1$ and $\rwb_2$ are optimized simultaneously.}
\label{fig20}
\efig

\smallskip\noindent{\bf PINN for forward models described by a PDE:} \\ 
For a more complex dynamical system with a PDE model such as $\dpdx{u(x,t)}{t}+\dpdx{u(x,t)}{x}=u(x,t)$. 
Here, the criterion must also account for the initial conditions (IC): $\left[u(x,0)-\hat{u}(x,0)\right]$, as well as the boundary conditions (BC): $\left[u(0,t)+\hat{u}(0,t)\right]$:
\beqnn
J(\rwb)&=&
\sum_{x,t} [u(x,t)-\hat{u}(x,t)]^2+
\sum_{x,t} \left[\dpdx{u(x,t)}{t}+\dpdx{u(x,t)}{x}-u(x,t)\right]^2 \\ 
& & +\sum_x \left[u(x,0)-\hat{u}(x,0)\right]^2
+\sum_t\left[u(0,t)+\hat{u}(0,t)\right]^2
\eeqnn
Figure~\ref{fig21} shows the scheme of this case. 

\bfig[h!]
\[
\barr{c}
x \\[12pt] t
\earr
\barr{c}
\NNsTwoTwoTwoOne
\earr
\hspace{-1.3cm}\rwb\hspace{1cm}
\hat{u}(x,t)
\Ra
\btabu{@{}c@{}}
u(x,t)\\ 
\includegraphics[width=2.5cm,height=2.5cm]{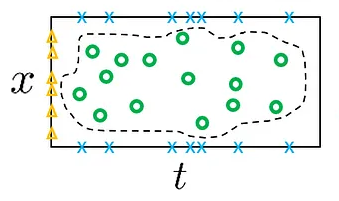}
\etabu
\Ra
\barr{@{}l@{}l}
\mbox{Residual error:}& \sum_{x,t} [u(x,t)-\hat{u}(x,t)]^2 \\
\mbox{Physics model error:}   & 
\disp{\sum_{x,t}} \left[\dpdx{u(x,t)}{t}+\dpdx{u(x,t)}{x}-u(x,t)\right]^2  \\
\mbox{Initial condition (IC):}& \sum_x \left[u(x,0)-\hat{u}(x,0)\right]^2 \\
\mbox{Boundary condition (BC):~}& \sum_t\left[u(0,t)+\hat{u}(0,t)\right]^2 \\
\earr
\]
\vspace{-12pt}
\caption{PINNfor forward models described by a PDE: Here, we have also account for IC and BC.}
\label{fig21}
\efig

\section{Applications}
We use these methods for developping Innovative Diagnostic and Preventive Maintenance Systems for industrial systems (Fans, Blowers, Turbine, Wind Turbine, etc) using Vibration, Acoustics, Infrared, and Visible images. 
The main classical and NNs methods we use are:
\bit 
\item Vibration analysis using Fourier and Wavelet analysis  
\item Acoustics: Sound Source localization and estimation using Acoustic imaging 
\item Infrared imaging: To monitor the temperature distribution
\item Visible images to monitor the system and its environment. 
\eit
Figure~\ref{fig22} shows some of the experimental measurement system we use.  
\bfig[h]
\[
\hspace*{-5mm}
\includegraphics[width=3.5cm,height=2cm]{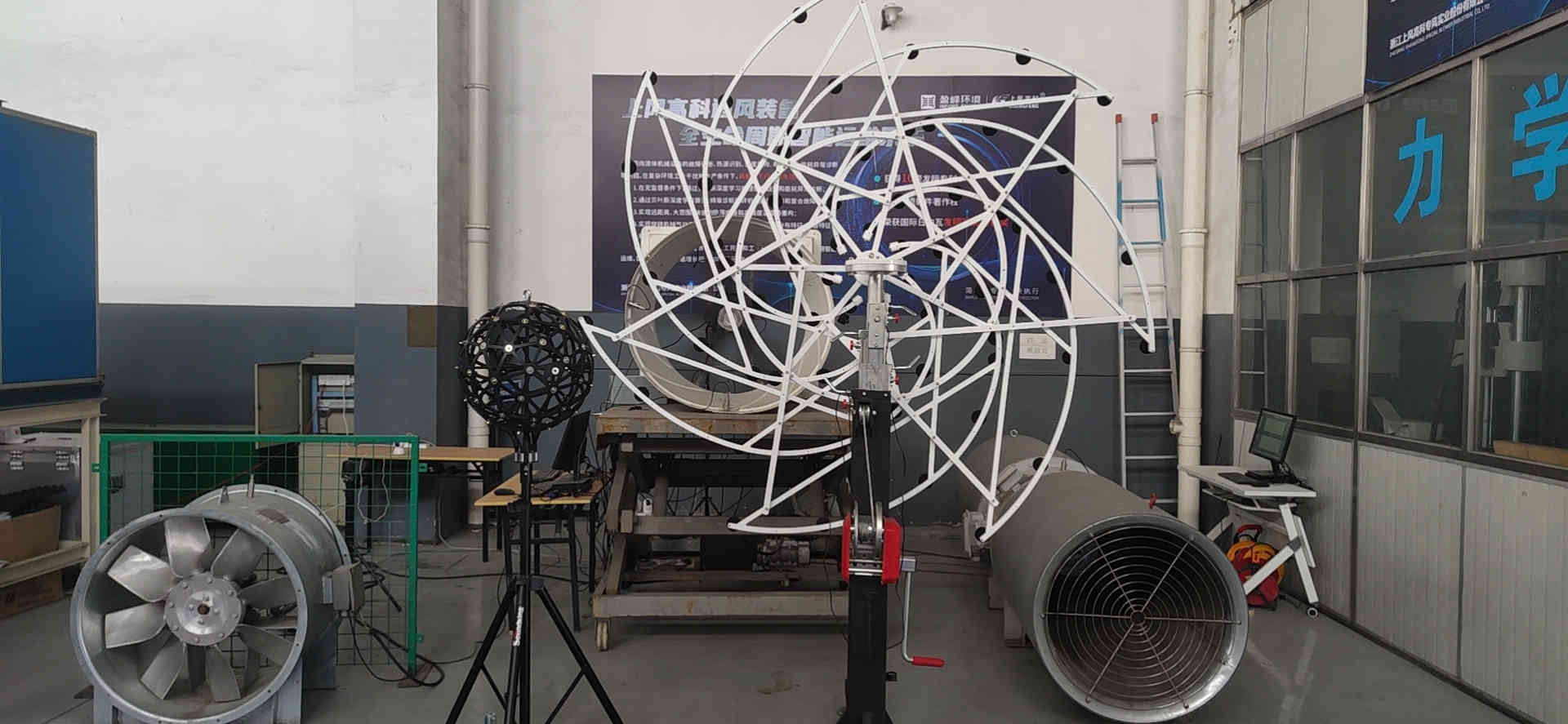}~
\includegraphics[width=3.5cm,height=2cm]{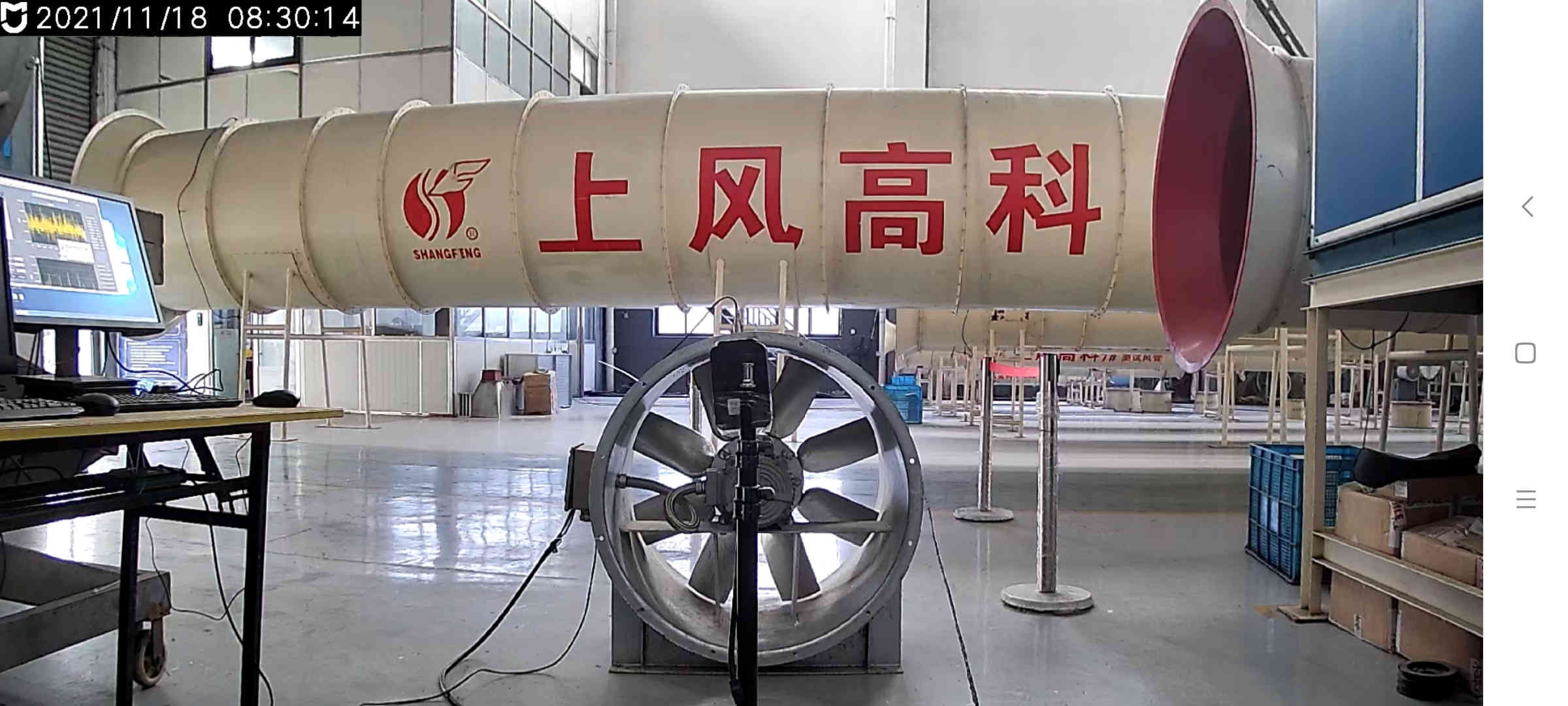}
\includegraphics[width=3.5cm,height=2cm]{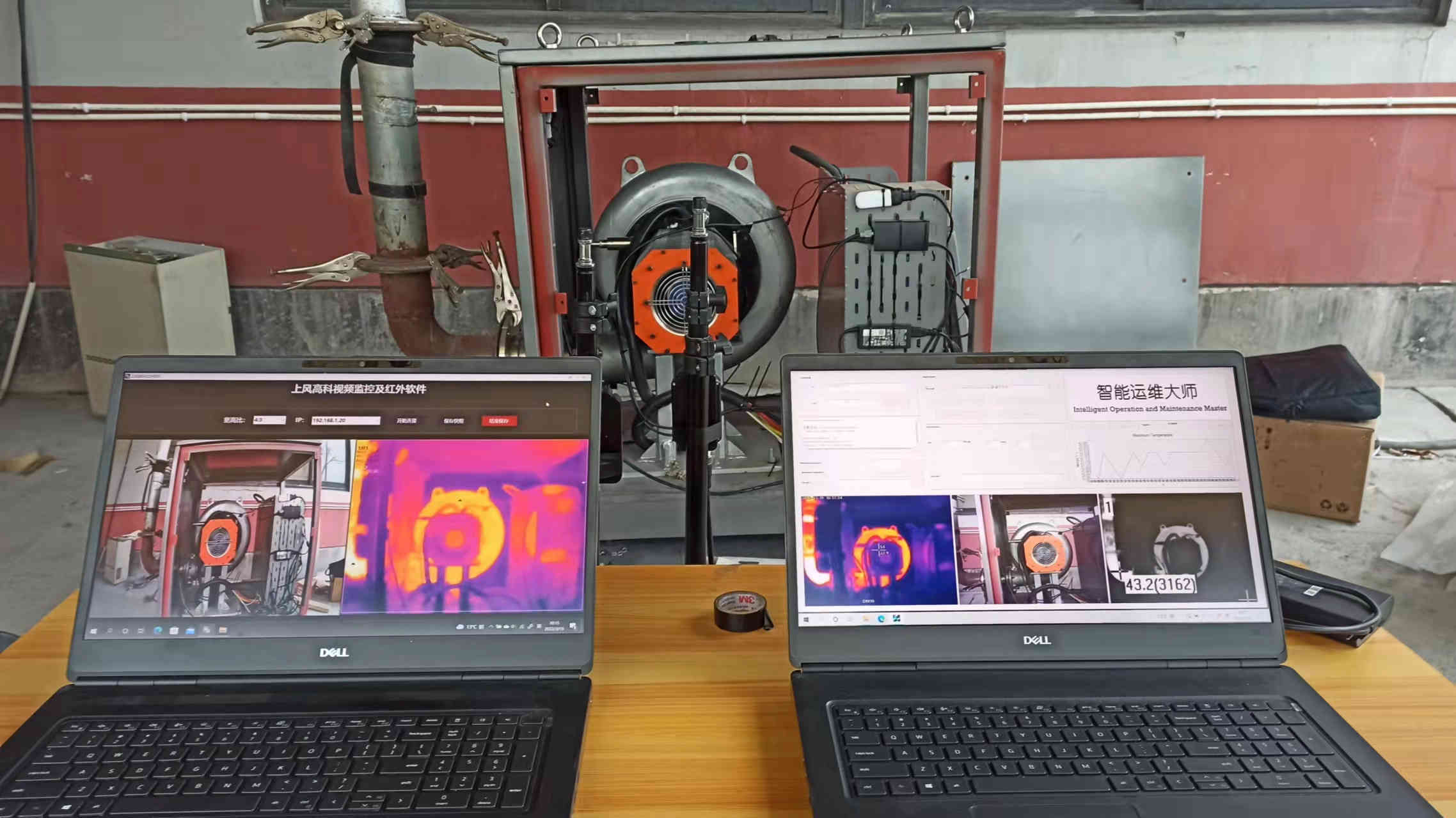}~
\includegraphics[width=3.5cm,height=2cm]{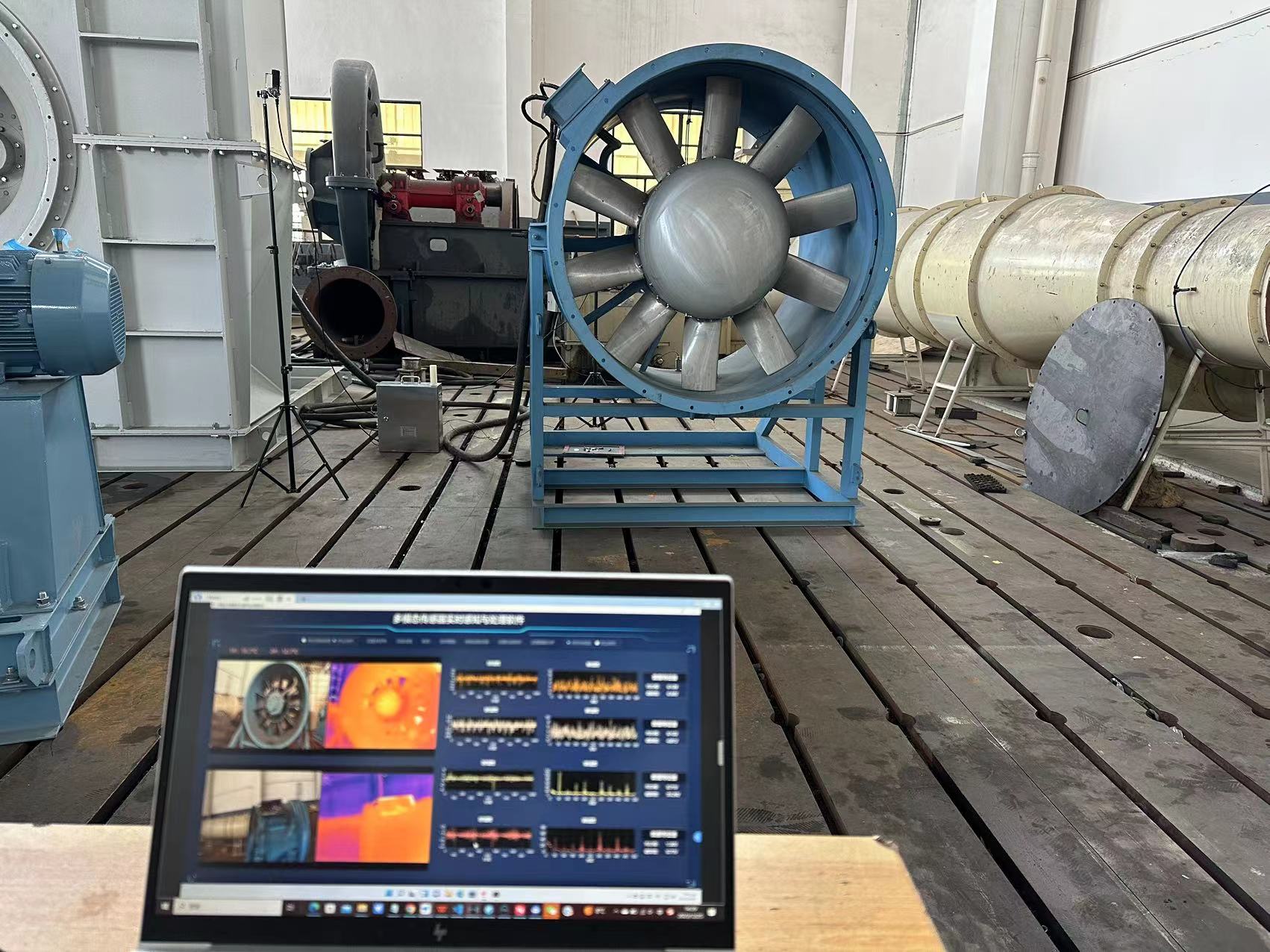}
\]
\vspace{-12pt}
\caption{Experimental measurement systems we use for Vibration, Acoustics, Infrared and Visual images for develpoing innovative diagnosis and preventive maintenance.}
\label{fig22}
\efig

\subsection{Infrared Image Processing}
Infrared images give a view of the temperature distribution, but they have low resolution and very noisy. There is a need for for denoising, deconvolution, super-resolution and segmentation and real temperature reconstruction, as shown in Figure~\ref{fig23}. 

\bfig[hbt]
\[
\barr{@{}c@{}c@{}c@{}l@{}c@{}c@{}}
Input & Denoising & Deconv & Segmentation & Temp. Recon. & Segmented\\
\mbox{IR image}&\ra\fbox{C1}-\fbox{Th}-\fbox{C2}\ra &
\fbox{C3}-\fbox{Thr}-\fbox{C4}\ra
& \fbox{SegNet}\ra 
&\fbox{Temp. Recons.}\ra &
\mbox{Temp. image}
\earr
\]
\vspace{-12pt}\caption{Infrared imaging: Denoising, Deconvolution, Segmentation and real temperature reconstruction.}
\label{fig23}
\efig
Some preliminary results using Deep Learning methods on simulated and real situations are given in Figure~\ref{fig24}.

\bfig[ht]
\bcc
\includegraphics[height=5cm,width=15cm]{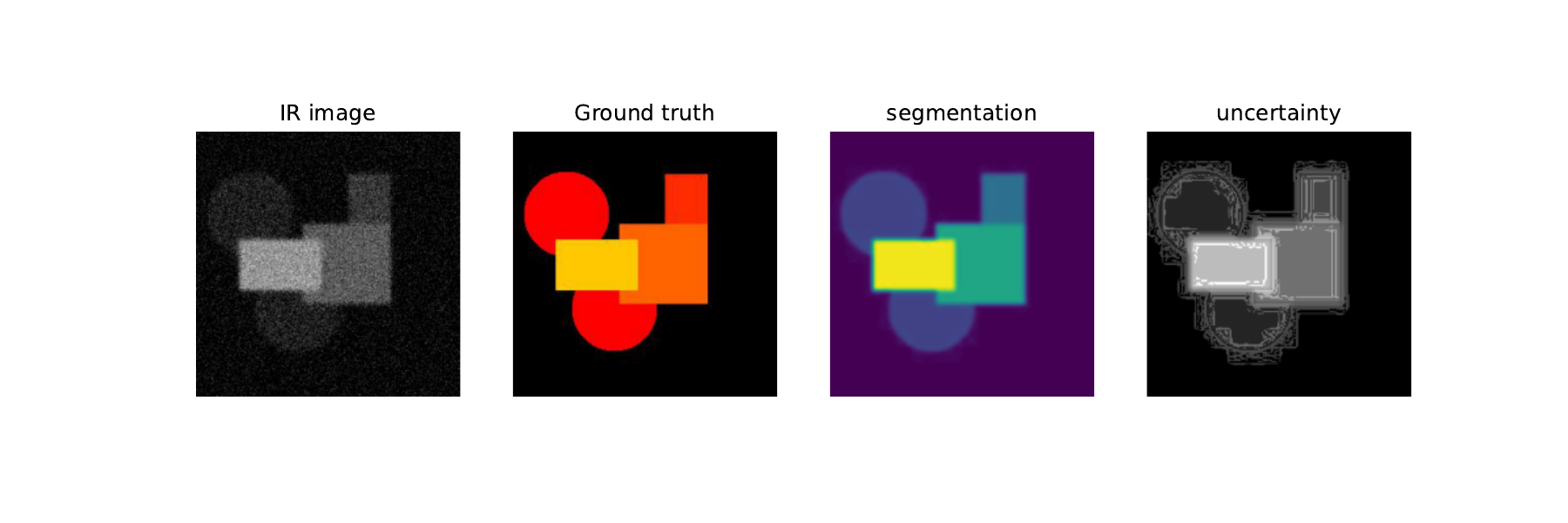}
\\[-48pt]
\includegraphics[height=5cm,width=15cm]{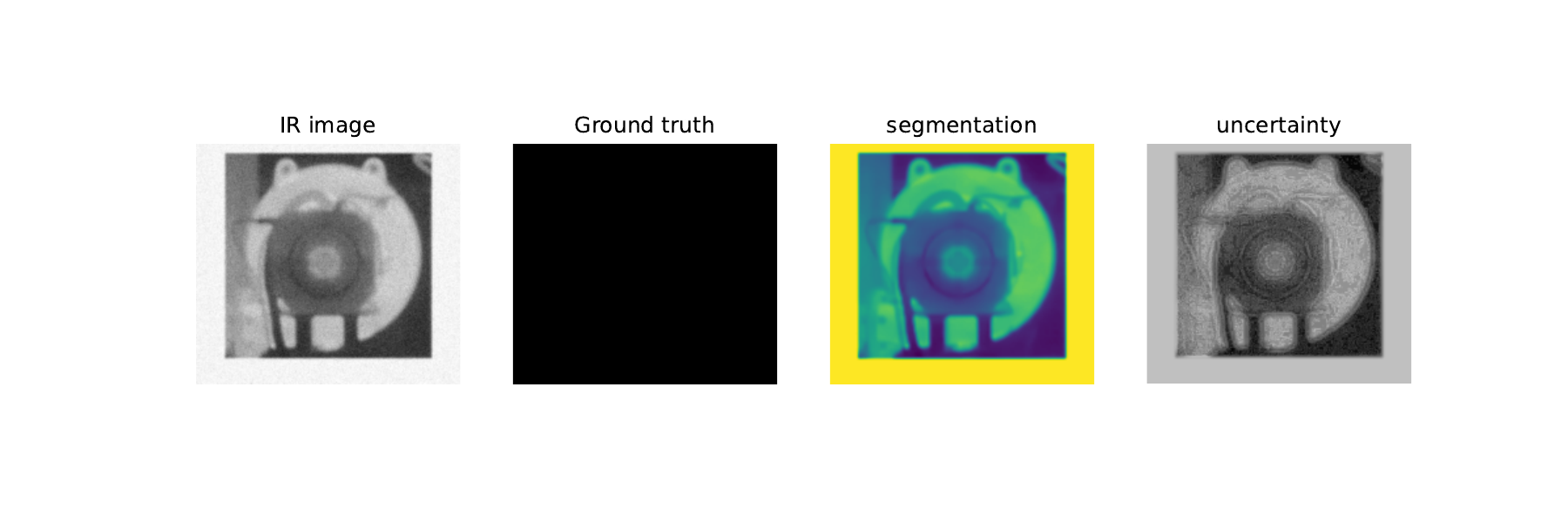}
\ecc
\vspace{-36pt}
\caption{One typical example of Bayesian Deep Learning method result obtained for infrared image segmentation: First raw shows the simulation results where we have a ground truth. Second raw is a real image where we do not have the ground truth.}
\label{fig24}
\efig

%PINN
%\cite{9152164,9084378,9000801,9049119,8434321,7468471},
%\cite{6467413,6933297,8081253,7953322,7471802}

%Applications

%\newpage
\section{Conclusions}
The main conclusions of this work are:
\bit
\item Deep Neural Networks (DNN) have been extendedly used in Computer Vision, and many other area. There are, nowadays, a great number of NN structures "on the shelf", and many users select one, and try to use. If it works, it is keeped, if not, just slects another, and so on!  
\item In particular, in Inverse problems of Computer Vision, there is a need for guding the selection of an explainable structure, and so, explainable ML and AI. 

\item In this work, a few directions have been proposed. These methods are classified as Model-based, Physics-based, or still, Physics Informed NN (PINN). 
Model based or Physics based models have become a necessity for all the Computer Vision and imaging systems to develop robust methods. 

\item PINN has originally been developed for inverse problems described by Ordinary or Partial Differentitial Equations (ODE/PDE). 
The main idea in all these problems is to choose appropriately NNs and use them as "Explainable" AI in industrial applications. 
\eit

%%%%%%%%%%%%%%%%%%%%%%%%%%%%%%%%%%%%%%%%%%
%\reftitle{References}

%\newpage
\def\bibdir{/home/sfgk/2024/Conferences/MaxEnt/biblio/}
\bibliography{MaxEnt2024_paper.bib,\bibdir infrared_imaging,\bibdir Acoustical_imaging,\bibdir NN_IP_Acoustic,\bibdir Chu_Ning,\bibdir PINN,\bibdir refs,\bibdir IP_ML_IEEE,\bibdir IEEEXplore2020,\bibdir amd_2022}
%\bibliography{MaxEnt2024_paper.bib,\bibdir Acoustical_imaging,\bibdir NN_IP_Acoustic}
\bibliographystyle{plain}

\end{document}

%% file: block_defs.tex
\def\SoftT{
\bpic(25,20)
  \put(0,10){\line(1,0){20}}
  \put(10,0){\line(0,1){20}}
  \put(15,10){\line(1,1){10}}
  \put(-5,0){\line(1,1){10}}
\epic
}

\def\Blocka{
\bpic(150,50)
  \put(0,20){\makebox(20,0){$\rfb^{(k)}$}}
  \put(20,20){\vector(1,0){10}}
  \put(30,10){\framebox(60,20){$(\Ib - \alpha \Hb^t\Hb)$}}
  \put(90,20){\vector(1,0){10}}
  \put(110,20){\circle{20}}
  \put(110,20){\makebox(0,0){$+$}}
  \put(120,20){\vector(1,0){10}}
  \put(130,10){\framebox(30,20){\SoftT}}
  \put(160,20){\vector(1,0){10}}
  \put(175,20){\makebox(20,0){$\rfb^{(k+1)}$}}
  \put(110,40){\vector(0,-1){10}}
  \put(105,45){\makebox(15,0){$\alpha\Hb^t\bgb$}}
\epic
}

\def\Blockwx#1#2{
\bpic(150,80)
  \put(20,20){\vector(1,0){10}}
  \put(30,10){\framebox(30,20){#2}}
  \put(60,20){\vector(1,0){10}}
  \put(80,20){\circle{20}}
  \put(80,20){\makebox(0,0){$+$}}
  \put(90,20){\vector(1,0){10}}
  \put(100,10){\framebox(30,20){\SoftT}}
  \put(130,20){\vector(1,0){10}}
  \put(80,40){\vector(0,-1){10}}
  \put(70,40){\framebox(20,15){#1}}
  \put(80,67){\vector(0,-1){10}}
  \put(75,70){\makebox(10,10){$\bgb$}}
  \put(80,75){\circle{20}}
\epic
}

\def\BlockDLw{\begin{picture}(100,100)
  \put(-160,10){\Blockwx{$\rWb_0$}{$\rWb^{(1)}$}}
  \put(-40,10){\Blockwx{$\rWb_0$}{$\rWb^{(2)}$}}
  \put(90,10){\Blockwx{$\rWb_0$}{$\rWb^{(K)}$}}
  \put(108,30){\makebox(0,0){...}}
  \put(-140,35){\makebox(0,0){$\rfbh^{(1)}$}}
  \put(245,35){\makebox(0,0){$\rfbh^{(K)}$}}
\end{picture}
}

%% file: NN_figures.tex
\def\NNsOneTwoTwoOne{
\begin{tikzpicture}[x=1cm, y=1cm, >=stealth]
\foreach \m/\l [count=\y] in {1}
  \node [circle,fill=black!20,minimum size=2mm] (input-\m) at (0,-\y-.5) {};
\foreach \m [count=\y] in {1,2}
  \node [circle,fill=black!30,minimum size=2mm] (hidden1-\m) at (.5,-\y) {};
\foreach \m [count=\y] in {1,2}
  \node [circle,fill=black!40,minimum size=2mm] (hidden2-\m) at (1,-\y) {};
\foreach \m [count=\y] in {1}
  \node [circle,fill=black!60,minimum size=2mm] (output-\m) at (1.5,-\y-.5) {};
\foreach \i in {1,...,1}
  \foreach \j in {1,...,2}
    \draw [->] (input-\i) -- (hidden1-\j);
\foreach \i in {1,...,2}
  \foreach \j in {1,...,2}
    \draw [->] (hidden1-\i) -- (hidden2-\j);
\foreach \i in {1,...,2}
  \foreach \j in {1}
    \draw [->] (hidden2-\i) -- (output-\j);
\end{tikzpicture}
}

\def\NNsTwoTwoTwoOne{
\begin{tikzpicture}[x=1cm, y=1cm, >=stealth]
\foreach \m/\l [count=\y] in {1,2}
  \node [circle,fill=black!20,minimum size=2mm] (input-\m) at (0,-\y) {};
\foreach \m [count=\y] in {1,2}
  \node [circle,fill=black!30,minimum size=2mm] (hidden1-\m) at (.5,-\y) {};
\foreach \m [count=\y] in {1,2}
  \node [circle,fill=black!40,minimum size=2mm] (hidden2-\m) at (1,-\y) {};
\foreach \m [count=\y] in {1}
  \node [circle,fill=black!60,minimum size=2mm] (output-\m) at (1.5,-\y-.5) {};
\foreach \i in {1,...,2}
  \foreach \j in {1,...,2}
    \draw [->] (input-\i) -- (hidden1-\j);
\foreach \i in {1,...,2}
  \foreach \j in {1,...,2}
    \draw [->] (hidden1-\i) -- (hidden2-\j);
\foreach \i in {1,...,2}
  \foreach \j in {1}
    \draw [->] (hidden2-\i) -- (output-\j);
\end{tikzpicture}
}